\documentclass[acmtog]{acmart}

\usepackage{adjustbox}
\usepackage{array}
\usepackage{gensymb}
\usepackage{multirow}

\usepackage{lipsum}

\usepackage[normalem]{ulem}
\useunder{\uline}{\ul}{}

\newcommand{\ua}{\uparrow}
\newcommand{\da}{\downarrow}

\usepackage{atbegshi}
\newcommand{\placetextbox}[3]{
  \setbox0=\hbox{#3}
  \AtBeginShipoutNext{\AtBeginShipoutUpperLeft{%
    \put(\dimexpr#1\paperwidth\relax,-\dimexpr#2\paperheight\relax)
    {\vtop{{\null}\makebox[0pt][c]{#3}}}%
  }}%
}

\AtBeginDocument{%
  \providecommand\BibTeX{{%
    \normalfont B\kern-0.5em{\scshape i\kern-0.25em b}\kern-0.8em\TeX}}}

\acmSubmissionID{1008}
\begin{teaserfigure}
  \includegraphics[width=\linewidth]{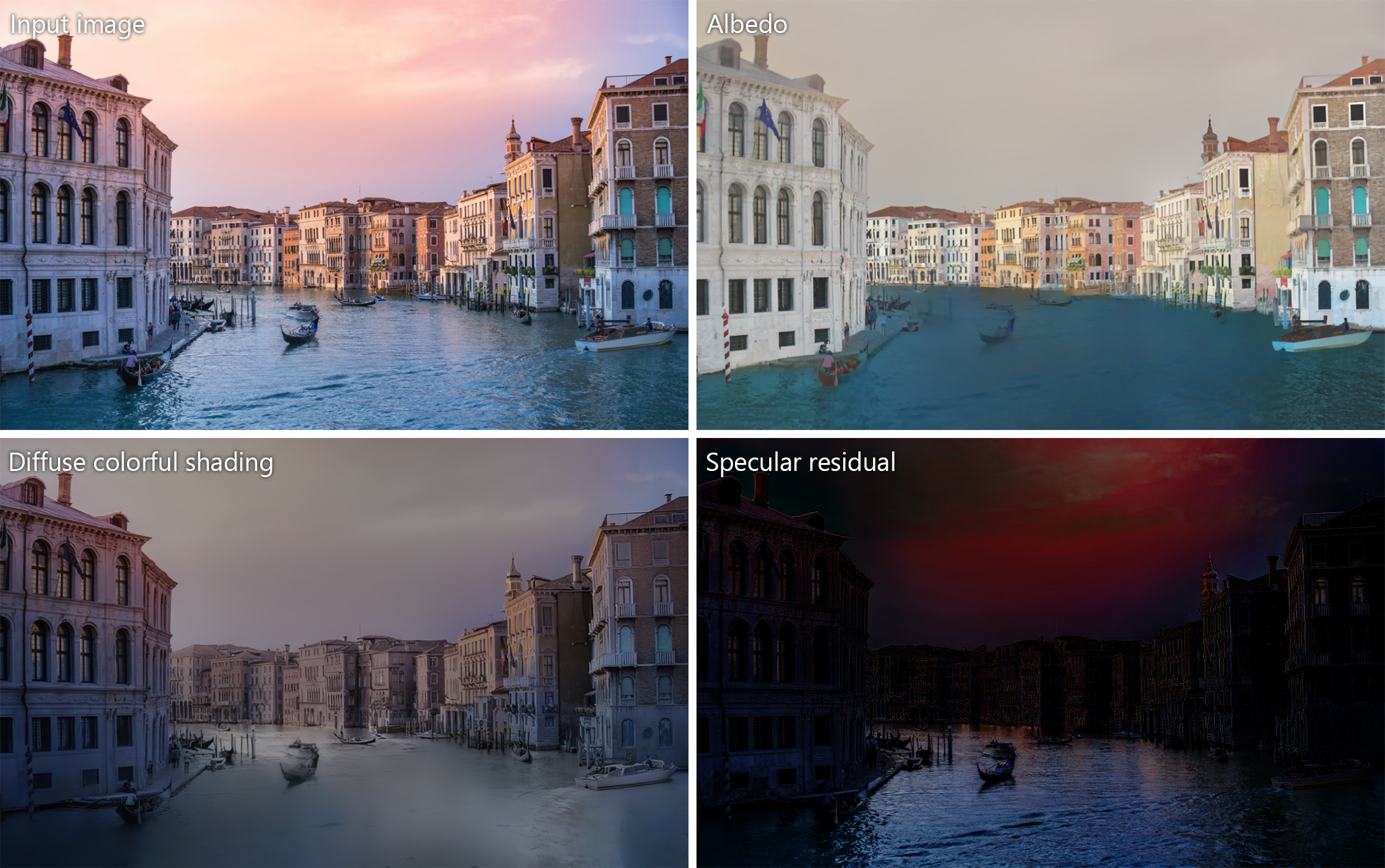}
  \caption{We present a method that can represent in-the-wild photographs in terms of albedo, diffuse shading, and non-diffuse residual components. Our shading layer reflects the colorful nature of multiple illuminants and secondary reflections in the scene, while our residual layer models the specularities and visible light sources. \hfill \footnotesize{Image from Unsplash by Rebe Adelaida.}}
  \label{fig:teaser}
\end{teaserfigure}

\begin{document}

\setcopyright{acmlicensed}
\acmJournal{TOG}
\acmYear{2024} \acmVolume{43} \acmNumber{6} \acmArticle{178} \acmMonth{12}\acmDOI{10.1145/3687984}

\title{Colorful Diffuse Intrinsic Image Decomposition in the Wild}

\author{Chris Careaga}
\author{Ya\u{g}{\i}z Aksoy}
\affiliation{
  \institution{Simon Fraser University}
  \city{Burnaby}
  \state{BC}
  \country{Canada}
}

\renewcommand{\shortauthors}{Careaga and Aksoy}

\begin{abstract}

Intrinsic image decomposition aims to separate the surface reflectance and the effects from the illumination given a single photograph. 
Due to the complexity of the problem, most prior works assume a single-color illumination and a Lambertian world, which limits their use in illumination-aware image editing applications. 
In this work, we separate an input image into its diffuse albedo, colorful diffuse shading, and specular residual components. 
We arrive at our result by gradually removing first the single-color illumination and then the Lambertian-world assumptions. 
We show that by dividing the problem into easier sub-problems, in-the-wild colorful diffuse shading estimation can be achieved despite the limited ground-truth datasets. 
Our extended intrinsic model enables illumination-aware analysis of photographs and can be used for image editing applications such as specularity removal and per-pixel white balancing.
\end{abstract}



\newcommand{\chris}[1]{\textcolor{purple}{{[chris: #1]}}}
\newcommand{\yagiz}[1]{\textcolor{red}{{[Ya\u{g}{\i}z: #1]}}}

\newcommand{\negvspace}{\vspace{-0.3cm}}

\maketitle

\placetextbox{0.14}{0.03}{\includegraphics[width=4cm]{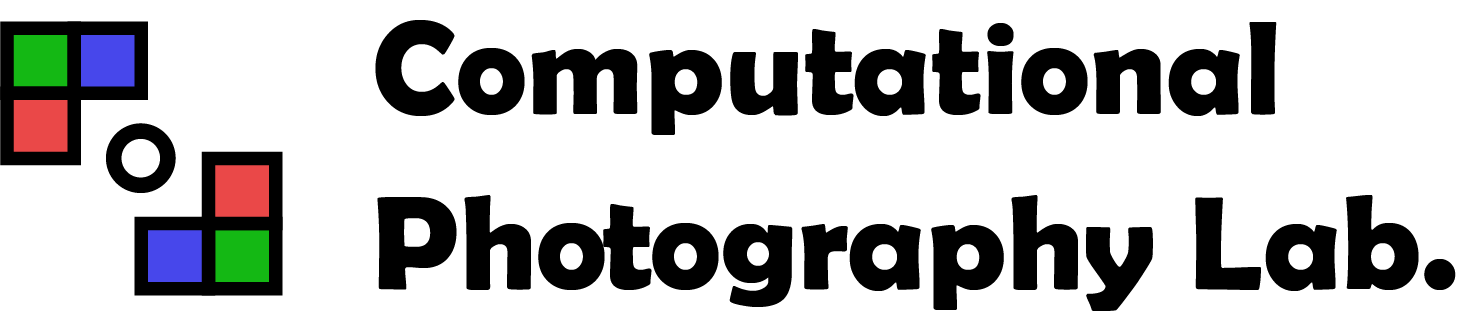}}
\placetextbox{0.85}{0.03}{Find the project web page here:}
\placetextbox{0.85}{0.045}{\textcolor{purple}{\url{https://yaksoy.github.io/ColorfulShading/}}}

\begin{figure*}
  \includegraphics[width=\linewidth]{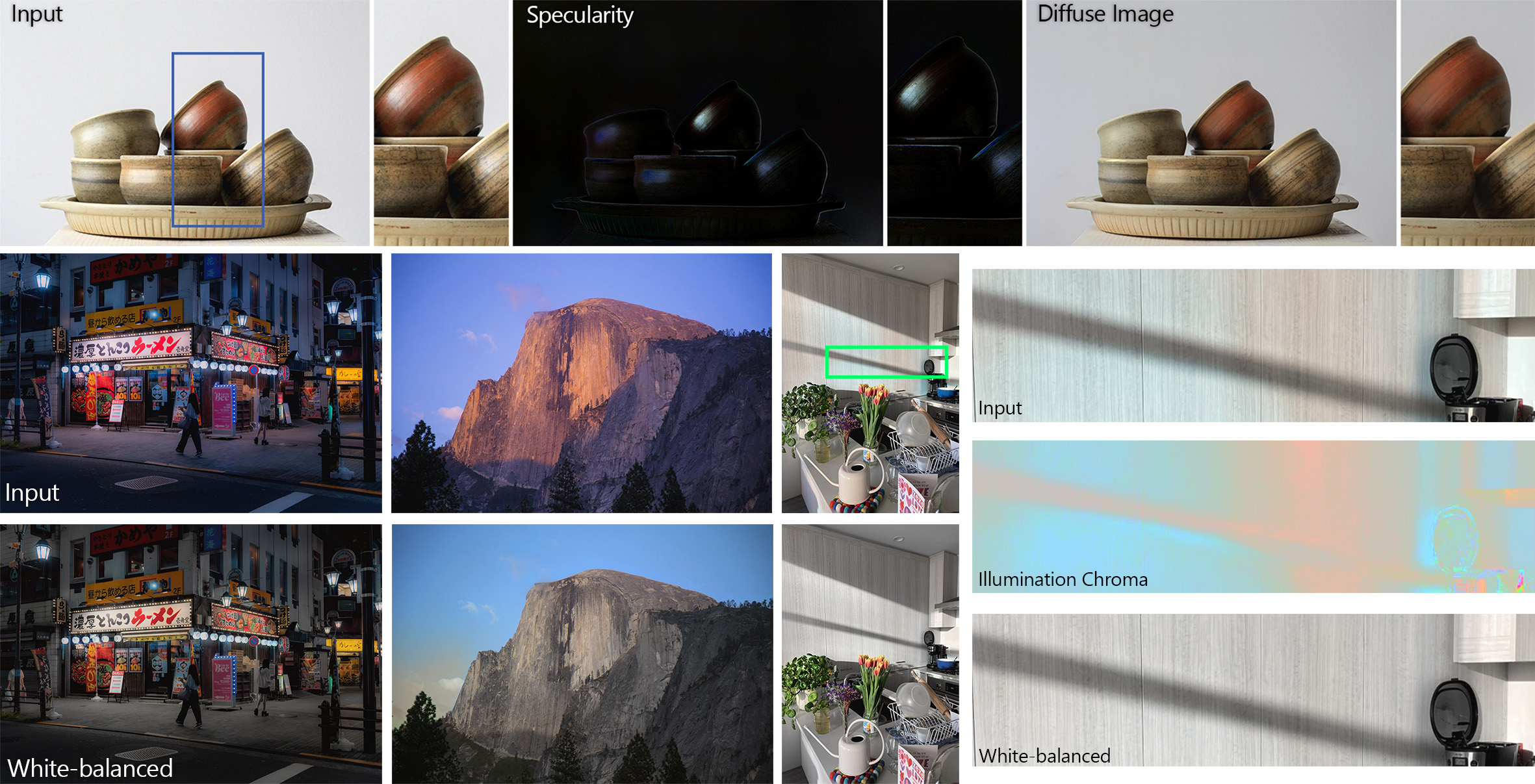}
  \caption{In this work, we extend the in-the-wild intrinsic decomposition formulations to include a colorful shading component as well as a non-diffuse residual component. This extended image formation enables illumination-aware image editing applications, such as specularity removal as shown at the top, and per-pixel white balancing. \hfill \footnotesize{Images from Unsplash by NorWood Themes (pots), mos design (street) and Josh Carter (mountain).}}
  \label{fig:intro_applications}
\end{figure*}

\section{Introduction}
\label{sec:intro}

Intrinsic image decomposition is a long-standing mid-level vision problem that aims to separate the surface reflectance and the effects of the illumination from a single photograph. 

Due to the complex interactions between the illumination and the geometry during image formation, it is a highly under-constrained task that requires high-level reasoning about the scene. 
The lack of real-world training data and the large domain gap between synthetic data and real-world photographs further complicate the task. 

Data-driven approaches to this problem have shown recent success, but prior works predominantly rely on the \textit{grayscale intrinsic diffuse model}, $I = A * S$,
where $I$ is the input image in linear RGB, $A$ is the 3-channel albedo, and $S$ is the single-channel grayscale shading. 
Although this model is shown to be useful in making the problem more tractable, it relies on two major assumptions that limit its applicability in real-world scenes. 

The first main assumption is the Lambertian world assumption that allows for the two-component multiplicative representation of the image by modeling all surfaces as diffuse. 
However, by ignoring specular surfaces, this model does not allow for separate editing of diffuse and non-diffuse illumination effects. 
The second assumption is the single-color shading that limits the model's representation of colorful illumination effects that are common in real scenes such as multiple light sources and inter-reflections.
This results in color effects being embedded in the albedo layer as shown in Figure~\ref{fig:albedo_steps}, limiting effectiveness in terms of color editing applications.

Few works in the literature attempt to further decompose illumination into diffuse shading and a non-diffuse residual, using the \textit{intrinsic residual model}, $I = A * S + R$, 
that extends the intrinsic diffuse model with an additive component $R$ that represents non-diffuse lighting effects such as specularities and visible light sources and defines $S$ as an RGB map that reflects the color of illumination. 
This enhanced capability to model real-world scenes comes at the cost of complexity, increasing the number of unknown variables from 4 to 9 per pixel, exacerbating the under-constrained nature of the problem. 
Coupled with a lack of diverse ground truth, prior methods that adopt the intrinsic residual model have been constrained to narrow contexts such as objects~\cite{shi2017shapenet, meka2018lime} or faces \cite{shah2023join, zhang2022phaced}.

\begin{figure*}
  \includegraphics[width=\linewidth]{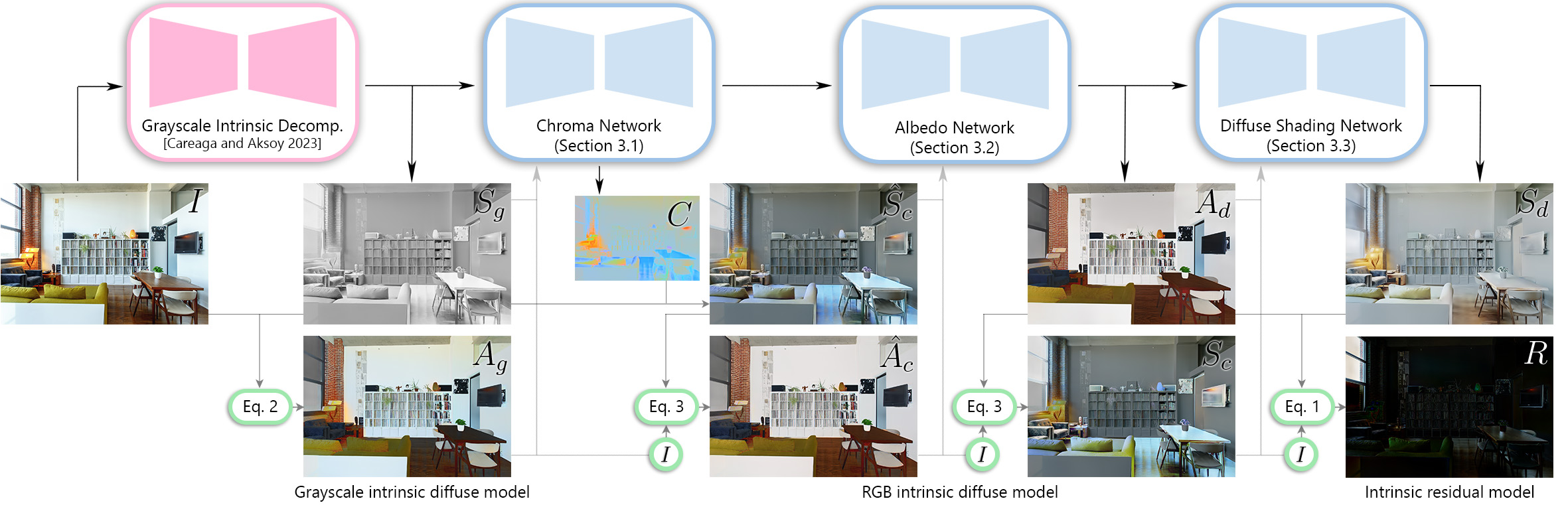}
  \caption{
    Our pipeline starts with an input image and a shading/albedo pair generated within the simplified grayscale intrinsic diffuse model generated via an off-the-shelf method. 
    We first extend the image formation model to include colorful shading, and estimate the shading color using our chroma network. 
    This color information is used as input in the second step where we estimate the high-resolution diffuse albedo. 
    In the final step, we remove the Lambertian-world assumption and estimate a colorful diffuse shading component and a non-diffuse residual layer.
    A single variable is estimated at each step, $S_g$, $C$, $A_d$, and $S_d$, respectively, and other intrinsic components are computed using the corresponding intrinsic image formation model with increasing representative power. \\\text{\ \ } \hfill \footnotesize{Image from Unsplash by Nathan Van Egmond.}
  }
  \label{fig:pipeline}
\end{figure*}

In this paper, we introduce a method that can generate decompositions under the intrinsic residual model for in-the-wild photographs. 
We start from a decomposition that uses the intrinsic diffuse model and gradually remove the single-color shading and the Lambertian world assumptions to estimate the diffuse albedo and the colorful diffuse shading at high resolutions. 
As summarized in Figure~\ref{fig:pipeline}, we first estimate the chroma of the shading using the global context present in the scene that is then used to create a sparse diffuse albedo. 
Given the diffuse albedo, we further decompose the shading into diffuse and specular components. 
We show that by breaking this highly under-constrained task into multiple conceptually simpler sub-problems, our method is able to generalize to complex in-the-wild scenes. 

We extensively evaluate our method's formulation and performance both qualitatively and quantitatively on common benchmarks as well as in-the-wild. 
We further demonstrate in Figure~\ref{fig:intro_applications} and Section~\ref{sec:applications} several illumination-aware image editing applications including per-pixel white-balancing and specularity removal that are made possible by the intrinsic residual model. 

\section{Related Work}
\label{sec:related}
Given the usefulness of intrinsic components in solving challenges in computational photography and image editing, the literature in this domain is extensive, covering multiple interrelated tasks. This section provides a summary of the field focusing on formulations and assumptions made by prior works in the context of our proposed approach. We refer the reader to the survey by~\citet{garces2022survey} for an in-depth discussion of the intrinsic decomposition literature.

\subsection{Intrinsic Decomposition}

\paragraph{Grayscale Diffuse Model}
The grayscale diffuse intrinsic model has been the predominant assumption since the earliest methods of intrinsic decomposition~\cite{tappen2005recovering, shen2008intrinsic}. 
This model has shown continued use due to its simplicity, creating a more tractable problem for both optimization-based~\cite{bell2014intrinsic, shen2011intrinsic, garces2012intrinsic, zhao2012closed} and data-driven~\cite{janner2017intrinsic, ma2018single, baslamisli2018cnn, baslamisli2018joint, li2018cgintrinsics, das2022pie, careaga2023intrinsic} approaches. As algorithms advance, this simplified model becomes more and more restrictive, causing inferred intrinsic components to stray further from physically accurate quantities.

\paragraph{RGB Diffuse Model}
Due to the shortcomings of the grayscale assumption, a few prior works explicitly model a colorful shading component. \citet{li2018learning} propose an unsupervised method for learning intrinsic components via time-lapse data, they parameterize their shading component as a grayscale map multiplied by a global RGB color cast. \citet{lettry2018unsupervised} propose a similar unsupervised training strategy but take it a step further by estimating an unconstrained RGB shading component. \citet{meka2021illumination} model an RGB shading layer and further decompose shading into separate light sources, but their method relies on low-level assumptions and user input, making it only suitable for simple scenes. Other works implicitly account for colorful shading effects by directly estimating albedo \cite{luo2020niid, luo2023crefnet, das2022pie} but these works typically constrain the albedo via an image reconstruction loss using the grayscale diffuse model, therefore lack in ability to accurately model colorful lighting effects.

\begin{figure*}
  \includegraphics[width=\linewidth]{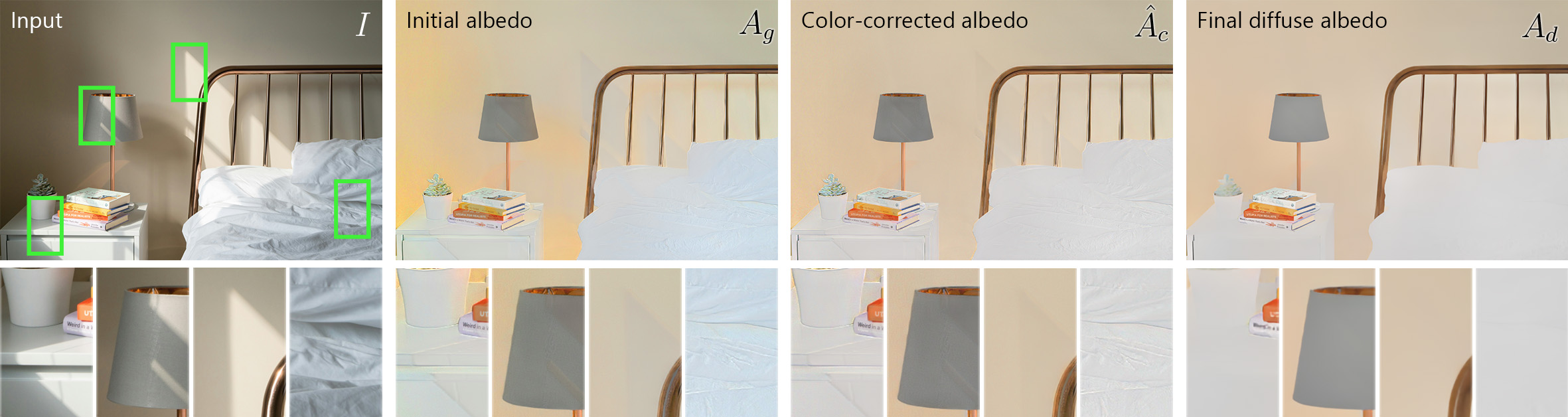}
  \caption{The initial albedo map that we use as input contains significant color shifts due to the grayscale shading assumption. Using the shading chroma estimated by our first network (Sec.~\ref{sec:method:chroma}), these color shifts are corrected but it fails to remove fine details coming from complex illumination. Our albedo estimation network (Sec.~\ref{sec:method:albedo}) is able to remove the effects of the illumination and estimate a sparse albedo map. \hfill \footnotesize{Image from Unsplash by Holly Stratton.} }
  \label{fig:albedo_steps}
\end{figure*}

\paragraph{Residual Model}
Extending beyond the well-known intrinsic diffuse model is not common in the literature. Given the difficulty of the problem and lack of real-world ground-truth supervision, prior works have only been able to estimate specularity in specific scenarios. \citet{shi2017shapenet} propose a method for estimating decompositions for singular objects, limiting the real-world applicability of their method. \citet{zhang2022phaced} use the residual model to estimate intrinsic components, but their method is specifically designed for human faces. \citet{shah2023join} also adopt the residual model. Although they evaluate their model on faces, material images, and simple scenes, they train separate networks for each task. 
\citet{kim2013specular} introduces an optimization formulation to infer the specularities without aiming full intrinsic decomposition. However, their low-level priors often lead to color edges being mislabeled.

Our method learns to estimate unconstrained RGB shading, both specular and diffuse, in the wild without the need for explicit assumptions or constraints. Despite being trained on indoor scenes, our diffuse shading network can generate accurate estimations for out-of-distribution images, as shown in Figure~\ref{fig:teaser}.

\subsection{Inverse Rendering}
Inverse rendering methods tackle the broader task of estimating all intrinsic scene parameters necessary to re-render an image. These methods typically estimate an albedo component explicitly and render shading via inferred geometry and an illumination model. Although this is a slightly different task formulation, inverse rendering methods are generally comparable to intrinsic decomposition methods as they still produce intrinsic components.

One of the earliest approaches by~\citet{barron2015sirfs} uses low-level priors to jointly recover scene intrinsics for simple scenes and isolated objects. \citet{karsch2014automatic} propose a method for indoor scenes that uses off-the-shelf albedo and depth estimation methods and infers illumination by optimizing for image reconstruction.
With the advancement of rendering capabilities, multiple data-driven methods have emerged~\cite{li2020inverse, sengupta19neural, zhu2022iris, zhu2022learning, wang2021learning, li2022panorama}. Given the limited availability of diverse training data, these methods focus on indoor scenes. 

Several recent works leverage diffusion-based image generation models to generate plausible intrinsic components conditioned on a given input image~\cite{zeng2024rgbx,kocsis2024iid,chen2024anything}. 
They model the problem as probabilistic, stemming from the under-constrained nature of the task. 
\citet{chen2024anything} focus on close-up object images and point to the ambiguity between the albedo and illumination colors. 
\citet{kocsis2024iid} focus on indoor images and point to different rendering engines and 3D models in CGI pipelines that occasionally embed several lighting effects in reflectance. 
They compensate for the random nature of their outputs by averaging over multiple estimations, which results in a loss of details. 
\citet{zeng2024rgbx} can directly estimate high-resolution intrinsic components, but suffer from being constrained to the latent space of the diffusion model as shown in Figure~\ref{fig:diffusion_comp}. 
Due to their fully generative modeling, these methods learn the appearance characteristics of the intrinsic components and work in a similar fashion to style transfer. 
\citet{zeng2024rgbx} make use of this aspect to demonstrate physically-guided image generation applications. 
In this work, we focus on the deterministic nature of real-world image formation and show that material and color ambiguities can be resolved through the context present in the scene.

\section{Method}
\label{sec:method}

\begin{figure*}
  \includegraphics[width=\linewidth]{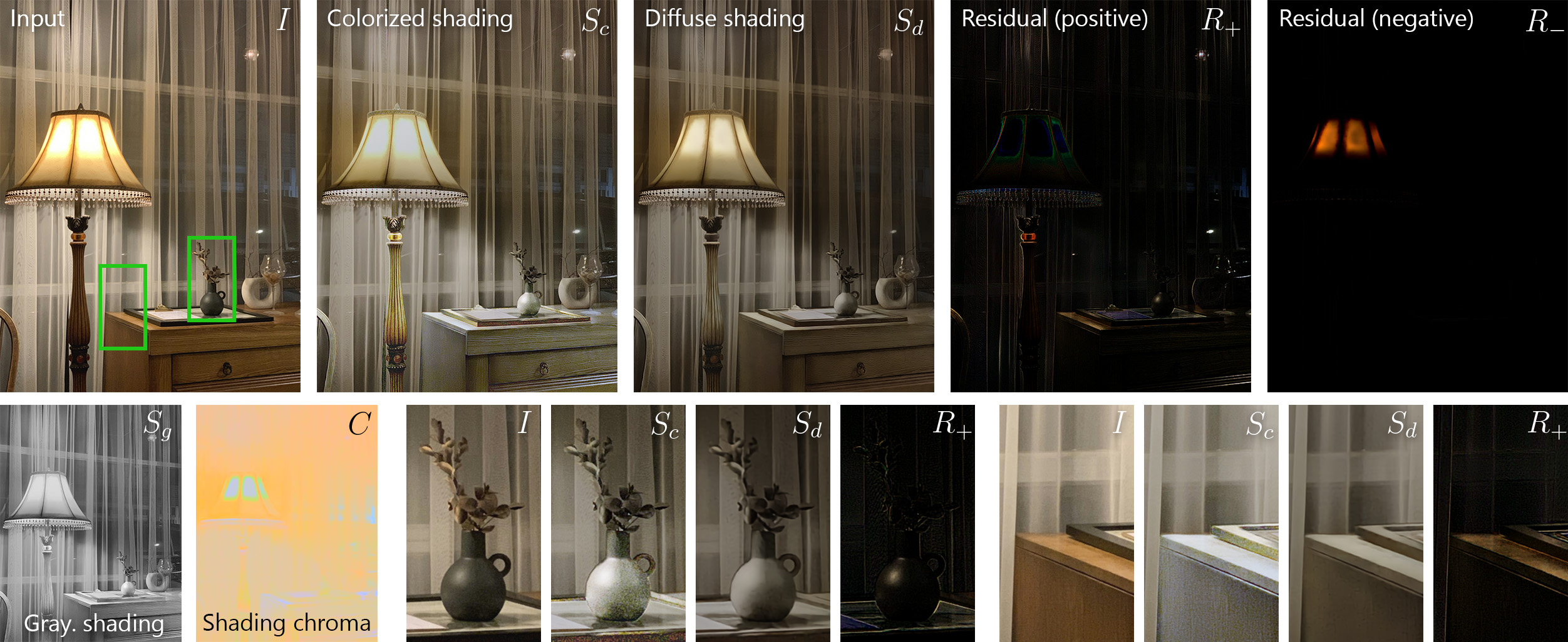}
  \caption{Starting from a grayscale shading estimation, we first estimate the shading chroma (Sec.~\ref{sec:method:chroma}) and create a colorized shading map. In the final step of our pipeline (Sec.~\ref{sec:method:shading}), we further separate the illumination into diffuse shading and non-diffuse residual components. The positive part of the residual represents the specularities in the scene, while the negative part shows the over-exposed regions. \hfill \footnotesize{Image from Unsplash by Jiwoo Park.}}
  \label{fig:shading_steps}
\end{figure*}

We aim to decompose an image $I$ into its diffuse albedo $A_d$ and colorful diffuse shading $S_d$ layers with a residual layer $R$ containing non-diffuse illumination effects using the intrinsic residual image formation model:
\begin{equation}
    I = A_d * S_d + R.
    \label{eq:intrinsic_residual}
\end{equation}
This highly under-constrained problem requires a network to reason about high-level contextual cues about scene geometry, global and local illumination conditions, and material properties.
The scarce high-resolution ground truth and the lack of real-world datasets for the diffuse shading component make it challenging for neural networks to statistically model the image formation in the wild. 

In order to achieve in-the-wild generalization, we divide the problem into simpler, physically-motivated sub-problems that are convenient for neural networks to model. 
We start from an existing intrinsic decomposition of the image that relies on the simplified Lambertian intrinsic model with a grayscale shading component $S_g$:
\begin{equation}
    I = A_g * S_g.
    \label{eq:intrinsic_diffuse_gray}
\end{equation}
We use the method by \citet{careaga2023intrinsic} to generate an $A_g$-$S_g$ pair that provides an initial starting point for our method.
We gradually remove the grayscale shading assumption, and then the Lambertian-world assumption, to arrive at our extended model in Equation~\ref{eq:intrinsic_residual}. 
Figure~\ref{fig:pipeline} gives an overview of our approach.

\subsection{Shading Chroma Estimation}
\label{sec:method:chroma}

One of the main reasons the grayscale shading assumption is adopted in the literature is that it significantly simplifies the problem by setting the albedo chromaticity to that of the input image. 
We begin our pipeline by abandoning the grayscale assumption and extend to the RGB intrinsic diffuse model:
\begin{equation}
    I = A_c * S_c,
    \label{eq:intrinsic_diffuse_rgb}
\end{equation}
that requires inferring the per-pixel chromaticity of the shading layer. 
For this purpose, we take the input grayscale shading $S_g$ as the luminance of $S_c$, and estimate the per-pixel chromaticities in our \emph{chroma network}. 
Borrowing ideas from color constancy literature \cite{murmann19multi, kim2021lsmi, barron2015constancy}, we define the chromaticity as color channel ratios:
\begin{equation}
    U = S^r_c / S^g_c \quad V = S^b_c/S^g_c. 
    \label{eq:chroma}
\end{equation}
Given that color channel ratios are unbounded variables, it is challenging to train neural networks with a direct loss on them. 
Hence, we use a simple mapping to the $[0-1]$ range following Careaga and Aksoy \shortcite{careaga2023intrinsic} and define our 2-channel target variable $C$:
\begin{equation}
    C = \biggl[ \frac{1}{U+1} , \frac{1}{V+1} \biggr].
    \label{eq:inv_chroma}
\end{equation}
Our chroma network takes the grayscale decomposition $(S_g, A_g)$ and the input image as a concatenated 7-channel input and estimates the 2-channel $C$. 
We train this network using the standard mean-squared error and the multi-scale gradient loss commonly utilized in the literature for mid-level vision tasks \cite{ranftl2020towards,li2018mega,li2018cgintrinsics,careaga2023intrinsic,miangolehSIDepth}:
\begin{subequations}
    \begin{gather}
        \mathcal{L}_{mse}(C) = \frac{1}{N} \sum_{i=1}^N (C_i - C^*_i)^2 \\
        \mathcal{L}_{msg}(C) = \frac{1}{NM} \sum_{i=1}^N \sum_{l=1}^M (\nabla C_{i,l} - \nabla C^*_{i,l})^2,
    \end{gather}
\end{subequations}
where $C^*$ is the ground-truth color component image, and $\nabla C_{i,l}$ is the gradient of $C$ at scale $l$. 

The shading chromaticity estimation requires an understanding of the global context present in the scene. 
It is also a low-frequency variable as discussed by \citet{lettry2018unsupervised}, making a low-resolution estimation viable. 
As a result, we utilize a convolutional architecture as detailed in Section~\ref{sec:method:network} and estimate $C$ at the receptive field-size resolution. 
We then combine this low-resolution $C$ with its luminance $S_g$ to construct the RGB shading layer $\hat{S}_c$.
Figure~\ref{fig:shading_steps} shows an example of our estimated chroma.

\begin{figure*}
  \includegraphics[width=\linewidth]{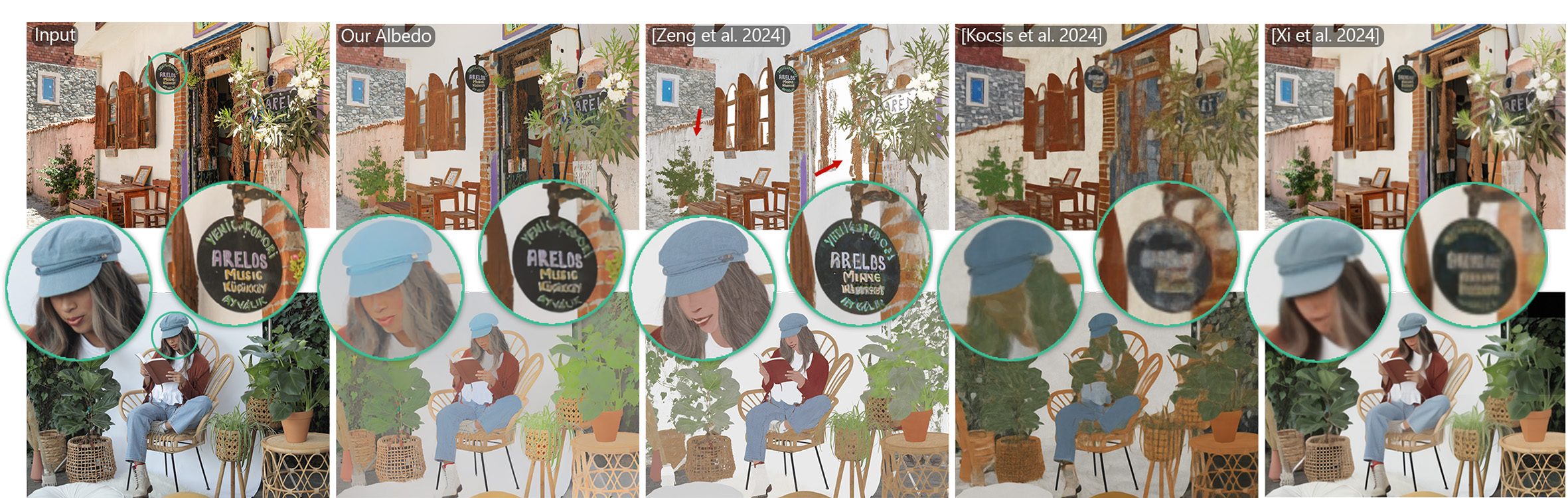}\\
  \vspace{2pt}
  \includegraphics[width=\linewidth]{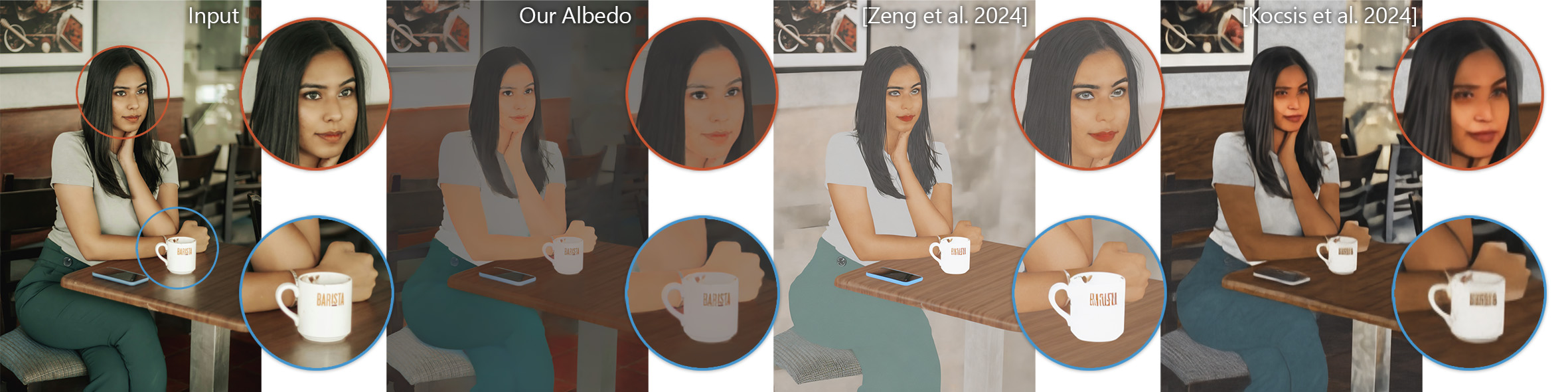} \\
  \includegraphics[width=\linewidth]{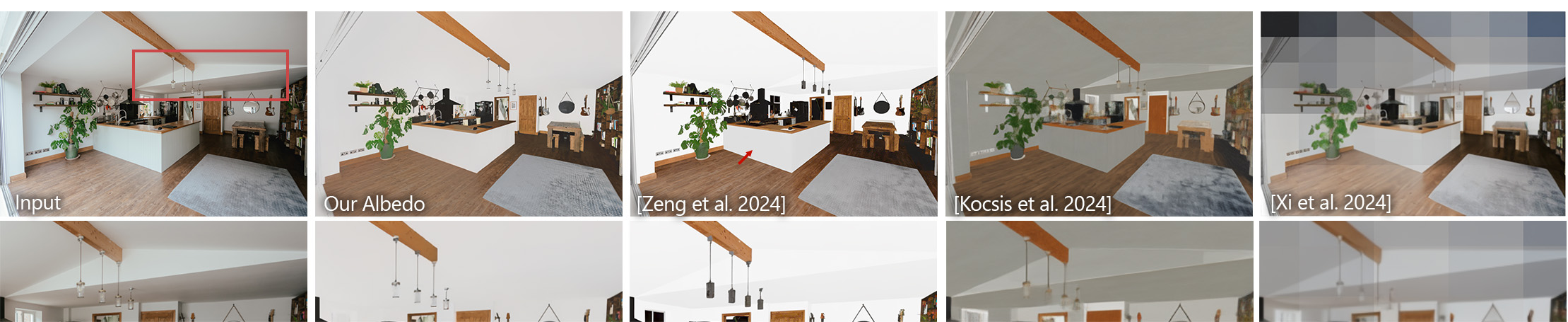}
  \caption{Given the large number of recently proposed diffusion-based methods, we provide a focused qualitative evaluation against these models. These examples show some of the shortcomings of utilizing generative modeling to address the problem of intrinsic decomposition. Since these methods learn a mapping in the latent space of large pre-trained generative models, their outputs can have unintended side-effects like warped faces, and illegible text. These alterations can have a negative impact on downstream editing applications. Additionally, although these methods can achieve the high-level appearance of albedo, they are highly dependent on their training data distribution which can cause effects such as large color shifts and baked-in shading. \\ 
  \text{\ } \hfill \footnotesize{Images from Unsplash (from top to bottom) by Mert Kahveci, Joel Muniz, Dollar Gill, and Annie Spratt} }
  \label{fig:diffusion_comp}
\end{figure*}

\begin{figure*}
  \includegraphics[width=\linewidth]{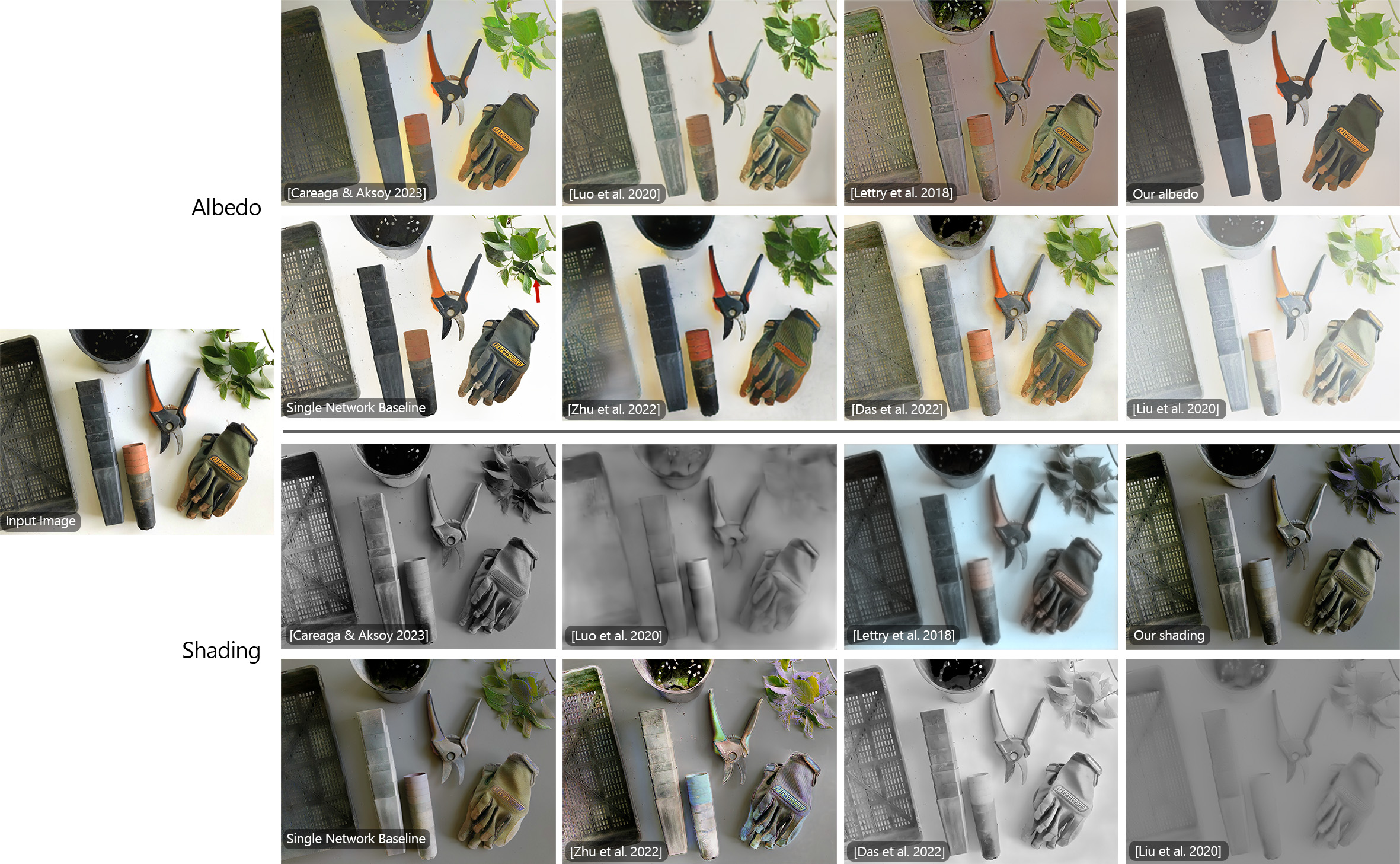}
  \caption{Prior grayscale shading works leave residual lighting such as interreflections in the albedo due to their formulation. Some works have attempted to model colorful lighting but the difficulty of the problem is exacerbated by the lack of ground truth data. Due to our formulation, our method is able to remove the colorful lighting effects from the albedo even for in-the-wild images. When compared to our single network baseline, our method generates a sparse albedo with colors that are true to the input image.\hfill \footnotesize{Image from Unsplash by Eco Warrior Princess.}}
  \label{fig:formulation_comp}
\end{figure*}

\subsection{Albedo Estimation}
\label{sec:method:albedo}

The albedo channel, when defined under the grayscale diffuse model, contains strong color shifts coming from colored illumination. 
The colorized shading $\hat{S}_c$ from our chroma network can be used to compute an approximation to the correct albedo, $\hat{A}_c$, using the RGB diffuse model in Equation~\ref{eq:intrinsic_diffuse_rgb}. 
However, due to the low-resolution chroma estimation and the lack of enforcement of sparse albedo values up to this point, $\hat{A}_c$ still exhibits illumination-related artifacts as Figure~\ref{fig:albedo_steps} demonstrates. 

In order to estimate our final diffuse albedo layer, we define our \emph{albedo network} that takes $\hat{A}_c$ and $\hat{S}_c$ as input together with the input image concatenated to be a 9-channel input and outputs the diffuse albedo $A_d$. 
With the global context on illumination color readily provided in its input, the task of our albedo network is to take advantage of the sparse nature of albedo and generate an accurate 3-channel diffuse albedo map. 
Similar to our chroma network, we use the mean-squared error $\mathcal{L}_{mse}(A)$ and the multi-scale gradient $\mathcal{L}_{msg}(A)$ losses on the albedo to train this network. 
As Figure~\ref{fig:albedo_steps} shows, this results in a flat albedo without illumination artifacts.

\subsubsection{Training datasets}

Intrinsic decomposition methods are typically trained with synthetic ground truth. 
Most synthetic intrinsic datasets readily provide the ground-truth albedo. 
Furthermore, real-world training data for albedo can be extracted from multi-illumination datasets \cite{careaga2023intrinsic}, greatly aiding the in-the-wild generalization. 
We train our chroma and albedo networks, as well as the ordinal network of \citet{careaga2023intrinsic}, using 8 synthetic datasets~\cite{roberts2021hypersim, zheng2020structure3d, zhu2022learning, krahenbuhl2018free, le2021eden, li2023matrixcity, yeh2022learning, wang2022intrinsic} and the multi-illumination dataset by \citet{murmann19multi} to provide a good variety of images during training, allowing our albedo estimation to generalize in-the-wild.

\subsection{Diffuse Shading Estimation}
\label{sec:method:shading}

With the diffuse albedo $A_d$ estimated, we are finally ready to abandon the Lambertian world assumption and estimate the colorful diffuse shading and non-diffuse illumination components in the intrinsic residual model in Equation~\ref{eq:intrinsic_residual}. 
Given that diffuse shading is highly correlated with the scene geometry, our \emph{diffuse shading network} needs to make use of the geometric cues in the scene to separate the diffuse effects from non-diffuse irradiance such as specularities and visible light sources. 
This problem can also be seen as the decomposition of $S_c = I / A_d$ in the RGB diffuse model in Equation~\ref{eq:intrinsic_diffuse_rgb} into diffuse and non-diffuse components.

Our diffuse shading network takes the diffuse albedo $A_d$, colorized shading from the diffuse model $S_c$, and the input image as a concatenated 9-channel input. 
We define the output in the inverse shading space \cite{careaga2023intrinsic} as a three-channel variable $D = 1 / (S_d+1)$ and use the mean-squared error $\mathcal{L}_{mse}(D)$ and the multi-scale gradient $\mathcal{L}_{msg}(D)$ losses during training.

Given the estimated diffuse shading $S_d$ and albedo $A_d$, we compute the residual non-diffuse layer using the intrinsic residual model in Equation~\ref{eq:intrinsic_residual}:
\begin{equation}
    R = I - (A_d * S_d).
\end{equation}
It should be noted that our estimated diffuse shading is unbounded, and therefore the diffuse image $(A_d * S_d)$ can exceed the input's $[0-1]$ range. 
This high-dynamic range property of our diffuse shading enables image enhancement applications as shown in Figure~\ref{sec:applications}. 
As a result of this property, our estimated residual has both negative and positive values. 
The positive part of the residual contains non-diffuse illumination effects such as specularities and visible light sources, while the negative residual shows over-exposed regions in the input image as shown in Figures~\ref{fig:shading_steps} and~\ref{fig:highlight_recovery}.

\subsubsection{Training dataset}

High-resolution synthetic datasets are scarce for diffuse shading and lack diversity, while real-world datasets are non-existent. 
This is the main reason why prior methods that focus on the residual model limit their application scenario to specific object classes. 
In our pipeline, however, our diffuse shading network readily gets the albedo and $S_c$ as input, which eases the contextual nature of its task. 
We train our diffuse network solely on the synthetic indoor Hypersim dataset \cite{roberts2021hypersim}. 
However, as our qualitative results demonstrate, our method generalizes to a wide range of in-the-wild images. 
This shows that by simplifying the task of each network, we are able to utilize the generalizability of our albedo estimation pipeline to achieve in-the-wild non-diffuse intrinsic decomposition.

\begin{figure}
  \includegraphics[width=\linewidth]{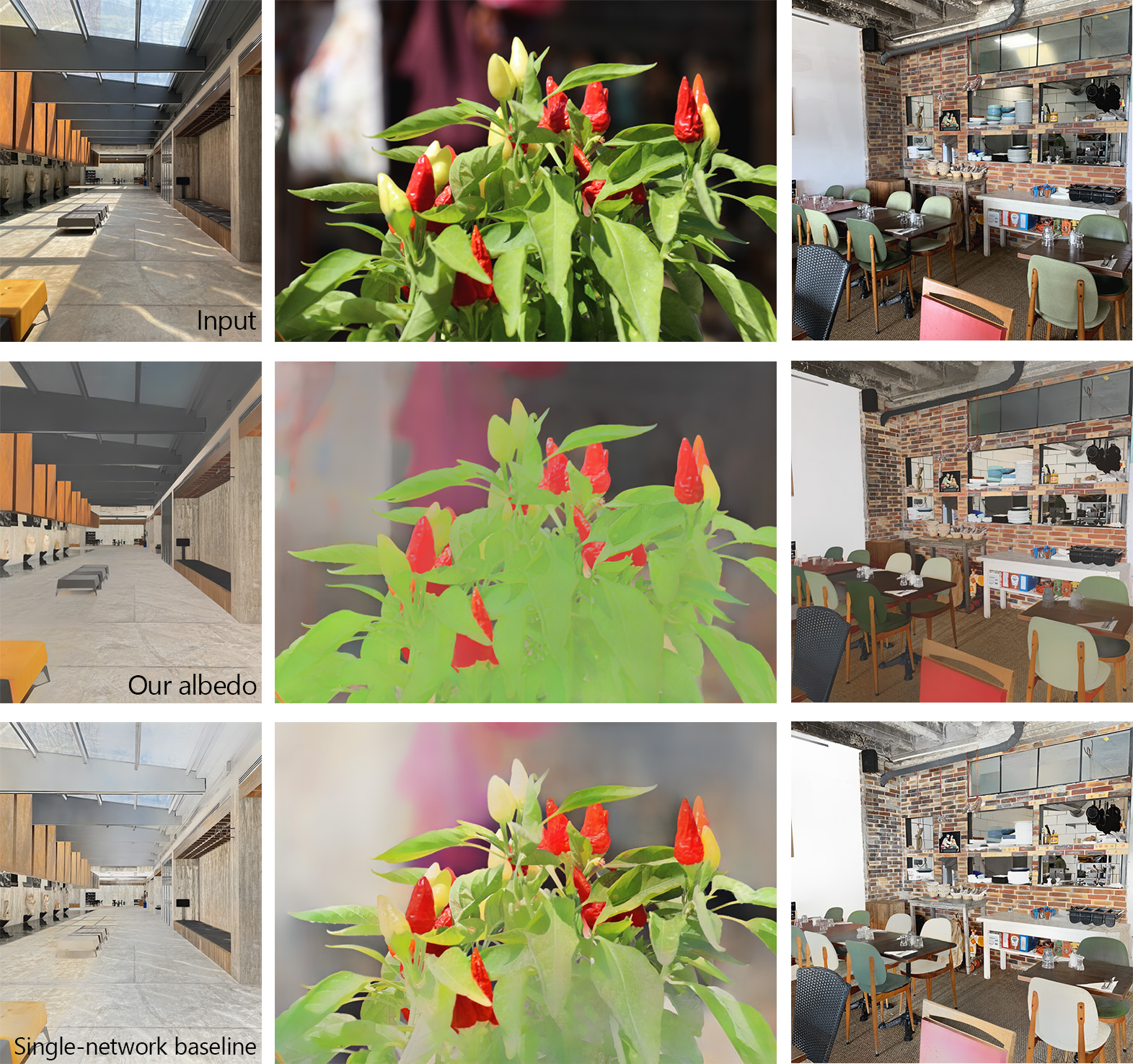}
  \caption{We train a large single network baseline on the same datasets as our full method (bottom row). Our pipeline achieves superior albedo estimation (middle row), especially on out-of-distribution high-resolution imagery. We attribute our performance to our multi-stage approach wherein each network accomplishes its simpler sub-task, learning generalizable behavior. \\ \text{\ }\hfill \footnotesize{Images from Unsplash by Alli Stefanova, Shalev Cohen, and Judith Girard-Marczak.}}
  \label{fig:single_net_comp}
\end{figure}

\subsection{Network Structure and Training}
\label{sec:method:network}
We utilize the same encoder-decoder architecture from~\cite{ranftl2020towards} that has been shown to be useful for various mid-level vision tasks for all of our networks. We use a sigmoid activation to output quantities strictly in the $[0-1]$ range. We train all the networks using the Adam optimizer with a learning rate of $10^{-5}$. Since intrinsic decomposition is an inherently scale-invariant task, typical formulations utilize scale-invariant losses when predicting intrinsic components. Due to the instability of these losses, we adopt the methodology of~\cite{careaga2023intrinsic} and set the arbitrary scale of ground-truth according to the input decomposition of each network. In doing these, regular loss functions can be used, training the networks to rely on the scale of the input to make their predictions. We provide further details in the supplementary.

\section{Experiments}
\label{sec:experiments}

We present quantitative evaluations of our method on common benchmarks, as well as qualitative comparisons to recent work. We extend our qualitative comparisons and show all the different components we estimate in a large set of in-the-wild images in the supplementary material. 

In order to show the effectiveness of our multi-stage pipeline, we compare our method to a single large model trained on the same datasets as a baseline. Specifically, we compare the albedo estimation of our approach (grayscale shading $\rightarrow$ shading chromaticity $\rightarrow$ albedo) to a single network that takes the input image and estimates albedo directly. The single network has 485 million parameters compared to our 4 networks which have 337 million parameters, cumulatively. We train the network for 1 million iterations with a batch size of 2 as that is the maximum size allowed by a 40GB GPU. We refer to this network as the "single-network baseline". It should be noted that such a single-network baseline is not practical for diffuse shading estimation due to the lack of real-world training datasets.

\subsection{Quantitative Evaluation}

Due to the lack of ground-truth benchmarks on diffuse shading, we report our quantitative analysis on the common test sets in the literature that focus on albedo estimation. 

\subsubsection{MAW Dataset}
Measured Albedo in the Wild (MAW) Dataset \cite{wu2023measured} has recently been introduced to measure real-world albedo accuracy in terms of intensity and color. 
The dataset consists of $\sim$850 indoor images and measured albedo within specific masked regions in the image. 
The albedo is measured using a known gray card placed on areas of homogeneous albedo. 
We focus our evaluation on two metrics that measure the accuracy of albedo intensity and chromaticity, respectively. 
The results are reported in Table~\ref{tab:measured_albedo}. 
As shown by the discrepancy between the intensity and chromaticity scores of the work by \citet{careaga2023intrinsic}, the grayscale shading assumption results in large inaccuracies in the color of the estimated albedo. 
Our initial shading chroma estimation is already able to compensate for these color shifts and scores the second-best in all metrics.
Our final refined albedo estimation further improves the results, outperforming all prior methods in terms of both intensity and chromaticity. When compared to the single-network baseline, our method achieves significantly higher performance, especially in terms of accurate albedo chromaticity. We attribute this improvement to our multi-stage approach which allows our method to generalize to the real-world images in MAW.

\begin{table}[]

\caption{Numerical results on the Measured Albedo in the Wild (MAW) Dataset~\cite{wu2023measured}. We achieve state-of-the-art performance on albedo estimation in terms of both intensity and chromaticity. Methods with an asterisk use the grayscale shading assumption and therefore have a fixed chromaticity score. For the first 7 methods we use the results computed by the authors of the MAW dataset.}
\begin{adjustbox}{width=0.85\linewidth}
{\renewcommand{\arraystretch}{1.2}}
\begin{tabular}{l c c}
\hline
Method & Intensity ($\times\ 100$)$\da$ & Chromaticity$\da$ \\ \hline
\citet{bell2014intrinsic} & 3.11 & 6.61 \\ 
\citet{li2018learning}  & 2.71 &  5.15 \\ 
\citet{li2018cgintrinsics} & 1.72 & 6.56* \\ 
\citet{sengupta19neural} & 2.17 & 6.39 \\ 
\citet{liu2020unsupervised} & 2.62 & 6.00 \\ 
\citet{li2020inverse} & 1.41 & 5.64 \\ 
\citet{luo2020niid} & 1.24 & 4.73 \\ 
\hline
\citet{lettry2018unsupervised} & 2.77 & 8.05 \\ 
\citet{zhu2022learning} & 1.44 & 4.94 \\ 
\citet{kocsis2024iid} & 1.13 & 5.35 \\ 
\citet{chen2024anything} & 0.98 & 4.12 \\ 
\citet{careaga2023intrinsic} & 0.57 & 6.56* \\ 
Single-Network Baseline & 0.69 & 4.15 \\
Ours ($\hat{A}_c$) & 0.56 & 3.50 \\
Ours & \textbf{0.54} & \textbf{3.37} \\ 

\hline

\end{tabular}
\end{adjustbox}
\label{tab:measured_albedo}
\end{table}

\begin{table}[]
\caption{Zero-shot albedo evaluation on the synthetic ARAP Dataset~\cite{bonneel2017intrinsic}. Our proposed method estimates the most accurate albedo across all zero-shot methods, even out-performing a non-zero-shot method marked with an asterisk in terms of SSIM.}
\begin{adjustbox}{width=0.8\linewidth}
{\renewcommand{\arraystretch}{1.2}}
\begin{tabular}{l|ccc}
Method                                      & LMSE$\da$     & RMSE$\da$       & SSIM$\ua$\\ \hline
Chromaticity                                & 0.038 & 0.193 & 0.710  \\
Constant Shading                            & 0.047 & 0.264 & 0.693  \\ 

\hline 
\citet{luo2020niid}*                         & 0.023 & 0.129 & 0.788  \\ \hline 
\citet{lettry2018unsupervised}               & 0.050 & 0.193 & 0.732  \\ 
\citet{kocsis2024iid}                        & 0.030 & 0.160 & 0.738  \\ 
\citet{zhu2022learning}                      & 0.029 & 0.184 & 0.729  \\ 
\citet{chen2024anything}                     & 0.038 & 0.171 & 0.692  \\ 
\citet{careaga2023intrinsic}                 & 0.035 & 0.162 & 0.751  \\ 
Single-Network Baseline           & 0.022 & 0.150 & \textbf{0.796}  \\ 
Ours ($\hat{A}_c$)                          & 0.025 & 0.156 & 0.752  \\ 
Ours                                        & \textbf{0.021} & \textbf{0.149} & \textbf{0.796}  \\  

\end{tabular}
\end{adjustbox}
\label{tab:arap}
\end{table}

\begin{figure*}
  \includegraphics[width=\linewidth]{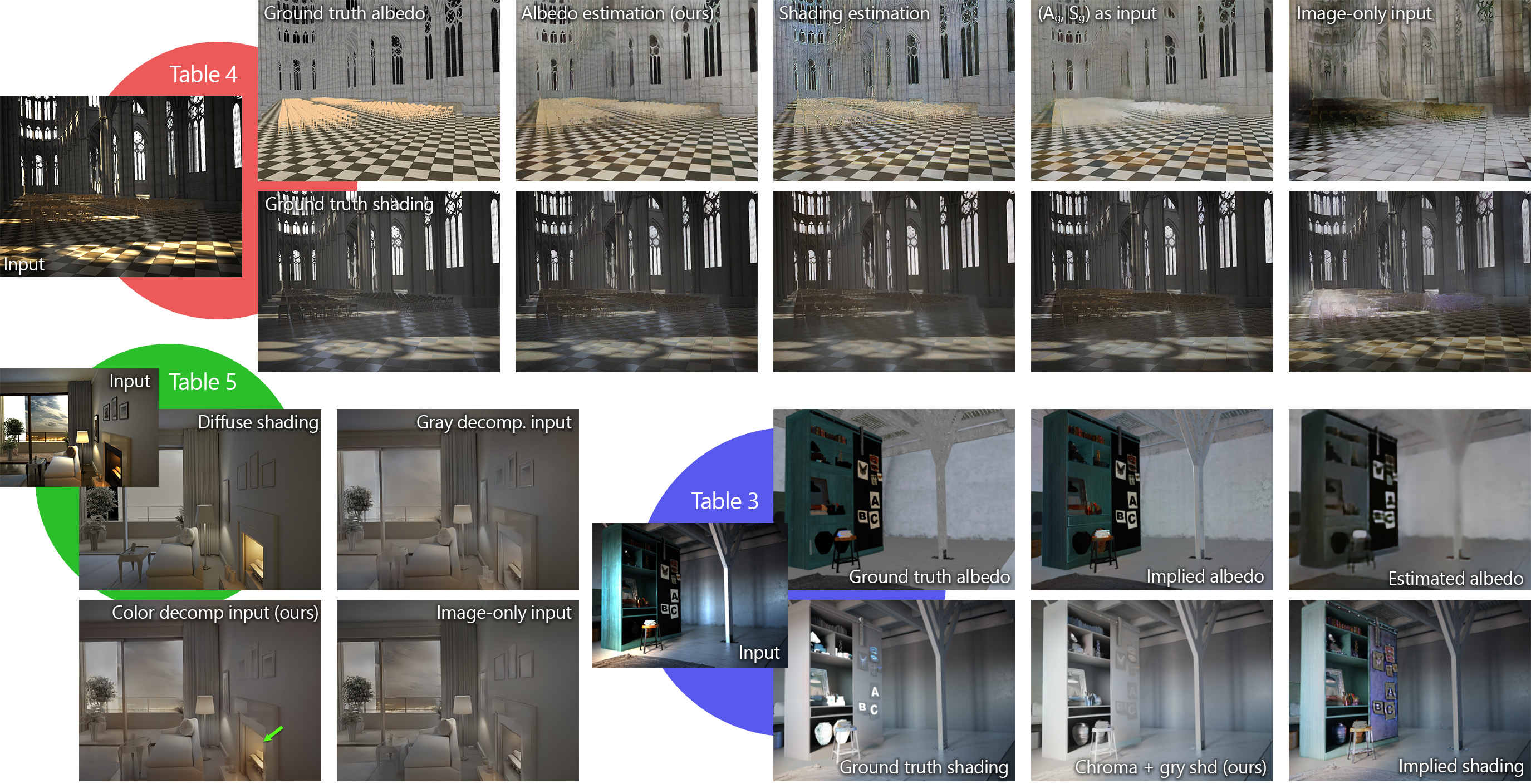}
  \caption{Qualitative examples from the Hypersim dataset for each of the three ablations discussed in Section~\ref{sec:experiments}. In Table~\ref{tab:color_abl} (bottom right) we show that estimating the chromaticity is a much easier task than estimating albedo directly, due to it's low-frequency nature. In Table~\ref{tab:albedo_abl} (top) we show that estimating albedo, given an initial colorful decomposition, is the best way to remove residual shading effects due to the inherent smoothness of the albedo component. In Table~\ref{tab:diffuse_abl} we show that our colorful decomposition is a vital input when estimating diffuse shading, and unlocks in-the-wild diffuse shading estimation.}
  \label{fig:ablations}
\end{figure*}

\subsubsection{ARAP Dataset}
In order to quantify the generalization abilities of each method to out-of-distribution scenes, we evaluate albedo estimation on the As Real as Possible (ARAP) Dataset~\cite{bonneel2017intrinsic}. The dataset consists of about 50 rendered scenes, from various sources. We augmented the dataset with three scenes from the MIST Dataset~\cite{hao2020mist} and also removed duplicated images and made sure each scene is equally represented in the dataset. We follow the same experimental setup as~\citet{careaga2023intrinsic} for computing evaluation metrics on the albedo.
The results are reported in Table~\ref{tab:arap}, with similar conclusions to Table~\ref{tab:measured_albedo}. We observe that in the rendered data setting of the ARAP dataset, the single-network baseline performs similarly to our network quantitatively despite its relatively poor performance on the MAW dataset. This shows that although the single network is able to accurately model the distribution of the synthetic training data, our multi-stage approach unlocks superior generalization to in-the-wild images.

\begin{figure*}
  \includegraphics[width=\linewidth]{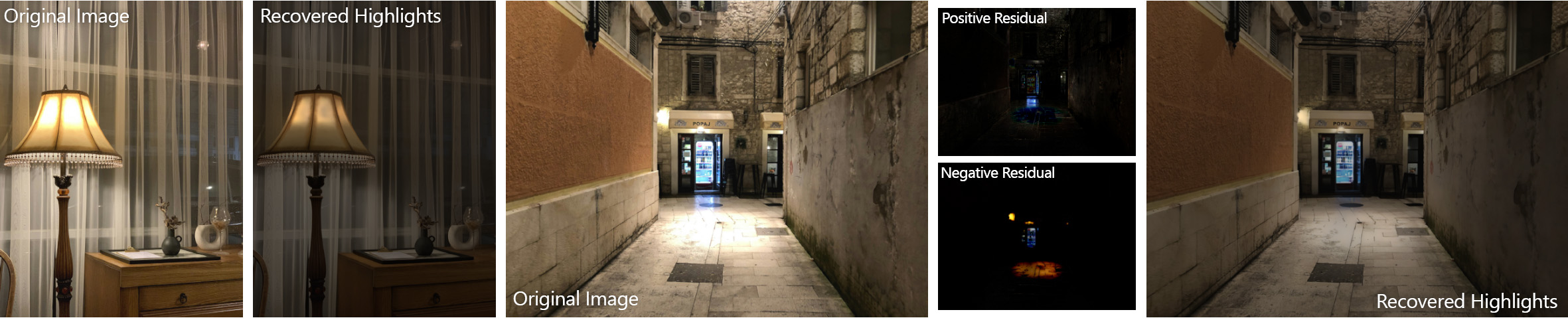}
  \caption{Since our estimated diffuse shading can represent unbounded light intensities, our method is able to recover information that was originally clipped in the input image. This clipped information results in negative values in our residual layer. \hfill \footnotesize{Image from Unsplash by Jiwoo Park}}
  \label{fig:highlight_recovery}
\end{figure*}

\subsection{Qualitative Evaluation}

Figure~\ref{fig:formulation_comp} show the results by \citet{careaga2023intrinsic}, \citet{luo2020niid}, ~\citet{das2022pie} and~\citet{liu2020unsupervised} which all adopt the grayscale intrinsic diffuse model, as well as the work by \citet{lettry2018unsupervised} that adopts the RGB intrinsic model in their unsupervised formulation. Additionally, we compare against~\citet{zhu2022learning} and our single-network baseline.
Since these two methods only provide albedo estimations, we compute shading by dividing the input image and estimated albedo.
When the grayscale model is enforced on the albedo-shading pair, the color of secondary illuminations creates color shifts in the albedo, as the results by \citet{careaga2023intrinsic} and \citet{das2022pie} show. 
Although the method of~\citet{zhu2022learning} estimates unconstrained albedo, their estimation still exhibits residual colors in the cast shadows. This color cast is removed in the result by \citet{luo2020niid}. 
However, since they still work within the grayscale model, their intrinsic components fail to faithfully reconstruct the image. The single-network baseline is able to remove the color cast from some shadows, but the network misses many shading effects (e.g. on the leaves), and generally shifts the colors of the input image (e.g. on the gloves).
We see a strong color cast and residual albedo colors in the shading by \citet{lettry2018unsupervised}, while our adoption of the intrinsic residual model allows us to estimate a clean albedo with colors of the secondary illuminations represented in our colorful shading.

We show in-the-wild comparisons against recent diffusion-based methods by \citet{zeng2024rgbx}, \citet{kocsis2024iid}, and \citet{chen2024anything} in Figure~\ref{fig:diffusion_comp}. 
The work by \citet{chen2024anything} suffers from low resolution in in-the-wild scenes, while in indoor scenes they are sometimes susceptible to a string tiling effect due to their high-resolution refinement. 
The work by \citet{kocsis2024iid}, on the other hand, struggles in out-of-distribution scenes and generates a low-resolution result due to the averaging of their results. 
The refinement and averaging strategies adopted by these methods result in $>10$ seconds run times, while our full pipeline takes around a second on average to generate a high-resolution result. 
The method by \citet{zeng2024rgbx} can generate sharp results but suffers from typical diffusion-based generation artifacts around text and may cause cartoonization of human faces. 
Our analytical modeling of the problem remains faithful to the input image and is able to generalize to out-of-distribution images effectively.

\begin{table}[]
\caption{Ablation experiment of our low-resolution chroma network. Our proposed method of estimating two-channel color information achieves much better results than directly estimating the albedo layer.}
\begin{adjustbox}{width=\linewidth}
{\renewcommand{\arraystretch}{1.2}}
\begin{tabular}{l|ccc|ccc}
\multirow{2}{*}{Method}        & \multicolumn{3}{c|}{Shading}                      & \multicolumn{3}{c}{Albedo}  \\
                               & LMSE$\da$       & RMSE$\da$       & SSIM$\ua$     & LMSE$\da$       & RMSE$\da$       & SSIM$\ua$    \\ \hline
direct albedo estimation       & 0.233           & 3.024           & 0.554          & 0.027          & 0.096          & 0.710           \\
ours - w/ chroma network       & \textbf{0.187}  & \textbf{2.866}  & \textbf{0.689} & \textbf{0.022}  & \textbf{0.083} & \textbf{0.815}  \\
\hline

\end{tabular}
\end{adjustbox}
\label{tab:color_abl}
\end{table}
\begin{table}[]
\caption{Ablation of albedo network formulation. By estimating the albedo given $\hat{S}_c$ our method yields better results than when directly estimating the shading at high resolution. This shows that the network can exploit the sparse nature of the albedo to make accurate predictions. Using any other inputs other than $\hat{S}_c$ and $\hat{A}_c$ results in decreased performance.}
\begin{adjustbox}{width=\linewidth}
{\renewcommand{\arraystretch}{1.2}}
\begin{tabular}{l|ccc|ccc}
\multirow{2}{*}{Method}                     & \multicolumn{3}{c|}{Shading}                     & \multicolumn{3}{c}{Albedo}  \\
                                            & LMSE$\da$       & RMSE$\da$       & SSIM$\ua$    & LMSE$\da$       & RMSE$\da$       & SSIM$\ua$    \\ \hline
image-only input                                 & 0.140          & 1.635          & 0.551           & 0.026          & 0.174          & 0.652            \\    
($S_g$, $A_g$) as input                     & 0.125          & 1.298          & 0.648           & 0.015          & 0.092          & 0.752            \\
shading estimation                          & \textbf{0.097} & \textbf{1.187} & 0.624           & 0.019          & 0.097          & 0.751            \\
ours - albedo estimation                    & 0.116          & \textbf{1.188} & \textbf{0.657}  & \textbf{0.012} & \textbf{0.090} & \textbf{0.757}   \\

\hline

\end{tabular}
\end{adjustbox}
\label{tab:albedo_abl}
\end{table}

\label{sec:experiments:ablations}
\subsection{Ablations}

In order to measure the performance impact of each individual design choice in our pipeline, we carry out multiple controlled experiments using the Hypersim dataset. For all ablations, we create a random scene split of the Hypersim dataset, with $66,000$ images for training and $6,000$ for evaluation. Each model variant is trained with a batch size of $8$ for $25,000$ iterations which is enough to give reasonable convergence given the similarity in the training and testing distributions.

\subsubsection{Chromaticity Estimation}
Table~\ref{tab:color_abl} shows the result of removing our chromaticity estimation formulation. The first row provides the scores for our proposed approach, estimating two color components and using the decomposition of~\citet{careaga2023intrinsic} as the luminance. The second row shows the result of directly estimating the albedo instead of using color components. Both networks receive the same input consisting of the input image and the grayscale decomposition from~\citet{careaga2023intrinsic}. The networks are evaluated at the receptive field size of the network (384px) in order to measure the global accuracy of each variant. We can see that by estimating the color components, the network learns the task much more effectively than when trying to directly reason about the albedo layer. This shows that this is an effective first step to estimating accurate albedo from the grayscale decomposition.

\begin{table}[]
\caption{Diffuse shading ablation experiment. When providing our diffuse shading network with diffuse albedo $A_d$ and the corresponding shading $S_c$ our network yields the best results. Any other input configuration results in worse performance, highlighting the effectiveness of our multi-step pipeline.}
\begin{adjustbox}{width=0.8\linewidth}
{\renewcommand{\arraystretch}{1.2}}
\begin{tabular}{l|ccc}

Method                                      & LMSE$\da$       & RMSE$\da$       & SSIM$\ua$   \\ \hline
image input                                 & 0.045 & 0.352 & 0.696 \\
grayscale decomp. input                     & 0.043 & 0.340 & 0.723 \\
ours - diffuse estimation                   & \textbf{0.040} & \textbf{0.329} & \textbf{0.728} \\
\hline

\end{tabular}
\end{adjustbox}
\label{tab:diffuse_abl}
\end{table}
\begin{figure*}
  \includegraphics[width=\linewidth]{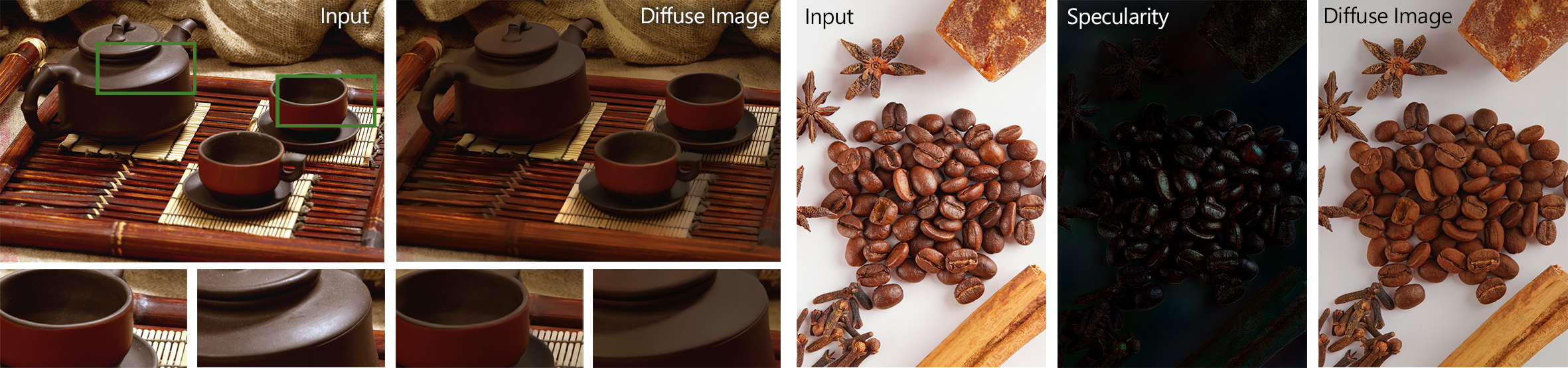}
  \caption{By removing the specular residual produced by our method, we are able to achieve specularity removal. Even though our diffuse shading network is solely trained on indoor images, it can still generate accurate estimations for diverse images. This capability is enabled by our multi-step formulation. \\ \text{\ }\hfill \footnotesize{Images from Unsplash by Kostiantyn Li (teapot) and Israel Albornoz (coffee).}}
  \label{fig:spec_removal}
\end{figure*}
\begin{figure*}
  \includegraphics[width=\linewidth]{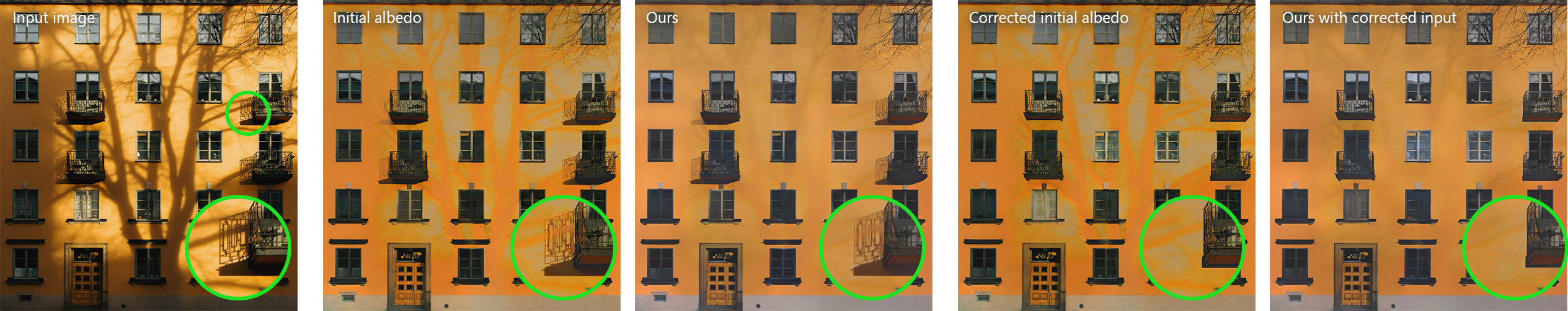}
  \caption{As our pipeline starts with a grayscale intrinsic decomposition as input, it may not be able to fix strong mistakes made by the initial model, such as the hard shadows of the balconies incorrectly included in the initial albedo on the left. However, we show that these mistakes can be fixed when the input albedo is corrected, in this case using Photoshop's content-aware inpainting tool. \hfill \footnotesize{Image from Unsplash by Jon Flobrant.}}
  \label{fig:limitations}
\end{figure*}

\subsubsection{Albedo Estimation}
Table~\ref{tab:albedo_abl} shows the result of alternatives to our second network that estimates high-resolution albedo. The first row shows our proposed approach of estimating the albedo given the decomposition with low-resolution chromaticity predicted by our first network. The second row shows that if we were to instead estimate high-resolution shading, our performance would drop in both albedo and shading estimation, showing that albedo estimation is a conceptually easier task for the network to model. The third and fourth rows show our albedo estimation network with different input configurations. Omitting the low-resolution chromaticity in the input shading results is a large performance decrease across all metrics, showing the value of estimating the albedo with our two-step approach. When completely omitting the intrinsic inputs and only providing the input image, the performance drops even more drastically, further showing the effectiveness of our multi-step approach.

\subsubsection{Diffuse Shading Estimation}
Table~\ref{tab:diffuse_abl} shows the impact of different input configurations on the diffuse shading network. The first row shows our proposed approach of using our estimated albedo and RGB shading layer as input. The second row shows that only providing the network with a grayscale decomposition, makes the task more difficult as the network has to reason about the shading color, resulting in lower scores on all metrics. Finally, the last row shows that our multi-step approach is essential for being able to estimate diffuse shading as trying to estimate it directly from the input image yields poor results. We present qualitative comparisons accompanying Tables~\ref{tab:color_abl}--\ref{tab:diffuse_abl} in Figure~\ref{fig:ablations}.

\section{Applications}
\label{sec:applications}

The intrinsic residual model allows for several computational photography applications by estimating a color component for the shading and separating diffuse and non-diffuse illumination effects. 
As demonstrated in Figures~\ref{fig:intro_applications} and~\ref{fig:spec_removal}, specularities in an image can be removed by computing the diffuse image $A_d*S_d$. 
Estimating the shading in color allows for per-pixel multi-illuminant white balancing, as shown in Figure~\ref{fig:intro_applications}. 
Our unbounded estimation of the diffuse shading allows us to recover details that are lost to clipping in the input image, as demonstrated in Figure~\ref{fig:highlight_recovery}.

\section{Limitations}
\label{sec:limitations}
Our method builds the intrinsic residual components by starting from the estimation of an existing method. 
While our networks are trained to account for erroneous initial estimations, they may also propagate some of the challenging mistakes as shown in Figure~\ref{fig:limitations}. 
In some simple cases, such mistakes can roughly be edited out in the input using commercial software, which in turn allows our pipeline to correct its estimation. 

\section{Conclusion and Future Work}
\label{sec:conclusion}
In this work, we present an intrinsic decomposition method that can successfully separate diffuse and non-diffuse lighting effects in the wild and at high resolutions. 
Our high-resolution performance and generalization ability come from our modeling of this highly under-constrained problem in physically-motivated sub-tasks. We demonstrate through quantitative analysis as well as qualitative examples that despite training our final diffuse network only on a synthetic indoor dataset, we are able to generalize to a wide variety of scenes including human faces and outdoor landscapes. We demonstrate new illumination-aware image editing applications that are made possible by adopting the intrinsic residual model.

We believe our method opens up multiple avenues for future work in this area. Our intrinsic residual model has the potential to improve intrinsics-based computational photography applications, some of which have been explored but could be improved by our approach, such as relighting \cite{careaga2023compositing}, flash photography \cite{Maralan2023Flash}, and HDR reconstruction \cite{dilleIntrinsicHDR}. Our method represents a large step towards developing physically accurate inverse rendering methods that generalize to in-the-wild images, and our components have the potential to be further decomposed into explicit lighting, BRDF parameters, and single- vs. multi-bounce contributions using more complex image formation models.

\begin{acks}
We would like to thank Zheng Zeng for promptly providing results for their method. We acknowledge the support of the Natural Sciences and Engineering Research Council of Canada (NSERC), [RGPIN-2020-05375]. 
\end{acks}

\newpage

\bibliographystyle{ACM-Reference-Format}
\bibliography{references}

\end{document}